\PassOptionsToPackage{table}{xcolor}
\documentclass[sigconf]{acmart}

\AtBeginDocument{%
  }

\setcopyright{acmlicensed}
\copyrightyear{2026}
\acmYear{2026}
\setcopyright{cc}
\setcctype{by}
\acmConference[MM '26] {Proceedings of the 34th ACM International Conference on Multimedia}{November 10--14, 2026}{Rio de Janeiro, Brazil.}
\acmBooktitle{Proceedings of the 34th ACM International Conference on Multimedia (MM '26), November 10--14, 2026, Rio de Janeiro, Brazil}
\acmISBN{979-8-4007-2213-4/2026/11}
\acmDOI{10.1145/3767308.3835550}
\usepackage{algorithm}
\usepackage{algorithmic}
\usepackage{multirow}
\usepackage[table]{xcolor}
\usepackage{booktabs}
\usepackage{amsmath,amsfonts}
\usepackage{makecell}
\usepackage{newfloat}
\usepackage{listings}

\begin{document}

\title{AnyTrack: Unifying Visual Object Tracking with Any Modalities}

\author{Hao Li}
\orcid{0009-0009-2668-7908}
\affiliation{%
  \institution{Army Engineering University of PLA}
  \city{Nanjing}
  \country{China}
}
\email{lihao@aeu.edu.cn}

\author{Yunzhi Zhuge}
\orcid{0000-0002-4288-4516}
\affiliation{%
  \institution{Dalian University of Technology}
  \city{Dalian}
  \country{China}
}
\email{zgyz@dlut.edu.cn}

\author{Wenning Hao}
\authornote{Corresponding authors.}
\orcid{0000-0002-1526-7889}
\affiliation{%
  \institution{Army Engineering University of PLA}
  \city{Nanjing}
  \country{China}
}
\email{hwnbox@aeu.edu.cn}

\author{Pingping Zhang}
\authornotemark[1]
\orcid{0000-0003-1206-1444}
\affiliation{%
  \institution{Dalian University of Technology}
  \city{Dalian}
  \country{China}
}
\email{zhpp@dlut.edu.cn}

\author{Xiaoxiong Zhang}
\orcid{0000-0002-3524-7543}
\affiliation{%
  \institution{National University of Defense Technology}
  \city{Nanjing}
  \country{China}
}
\email{xiaoxiongzhang@nudt.edu.cn}

\author{Dong Wang}
\orcid{0000-0002-6976-4004}
\affiliation{%
  \institution{Dalian University of Technology}
  \city{Dalian}
  \country{China}
}
\email{wdice@dlut.edu.cn}

\author{Huchuan Lu}
\orcid{0000-0002-6668-9758}
\affiliation{%
  \institution{Dalian University of Technology}
  \city{Dalian}
  \country{China}
}
\email{lhchuan@dlut.edu.cn}

\renewcommand{\shortauthors}{Li et al.}

\begin{abstract}
Visual object tracking aims to continuously locate specific targets within sequential frames, evolving from single-modal methods to multi-modal ones.
However, existing multi-modal trackers are typically designed for fixed modality combinations, requiring separate models for different inputs.
This leads to a poor adaptability to missing or imperfect modalities, and limited generalization.
To address these issues, we propose a novel unified framework called AnyTrack for object tracking with any modalities.
Specifically, we design a Modality-aware Interaction Module (MIM) to facilitate dynamic interaction across diverse modalities.
This module bridges modality discrepancies and aggregates temporal cues to maintain spatio-temporal consistency during cross-modal interaction.
Furthermore, we introduce a Context Understanding Module (CUM) to establish spatial correspondence between visual features and target locations via global-local prompts.
This module employs target-aware context modeling to enhance foreground-background discrimination for precise localization.
Finally, to support the training and evaluation under diverse modalities, we extend existing multi-modal object tracking benchmarks by incorporating grayscale images, language descriptions, and audio clips.
Extensive experiments with both complete and missing modality settings demonstrate that our AnyTrack achieves state-of-the-art performance, validating its effectiveness and flexibility.
The source code is available at \href{https://github.com/IdolLab/AnyTrack}{https://github.com/IdolLab/AnyTrack}.
\end{abstract}



\begin{teaserfigure}
  \includegraphics[width=\textwidth]{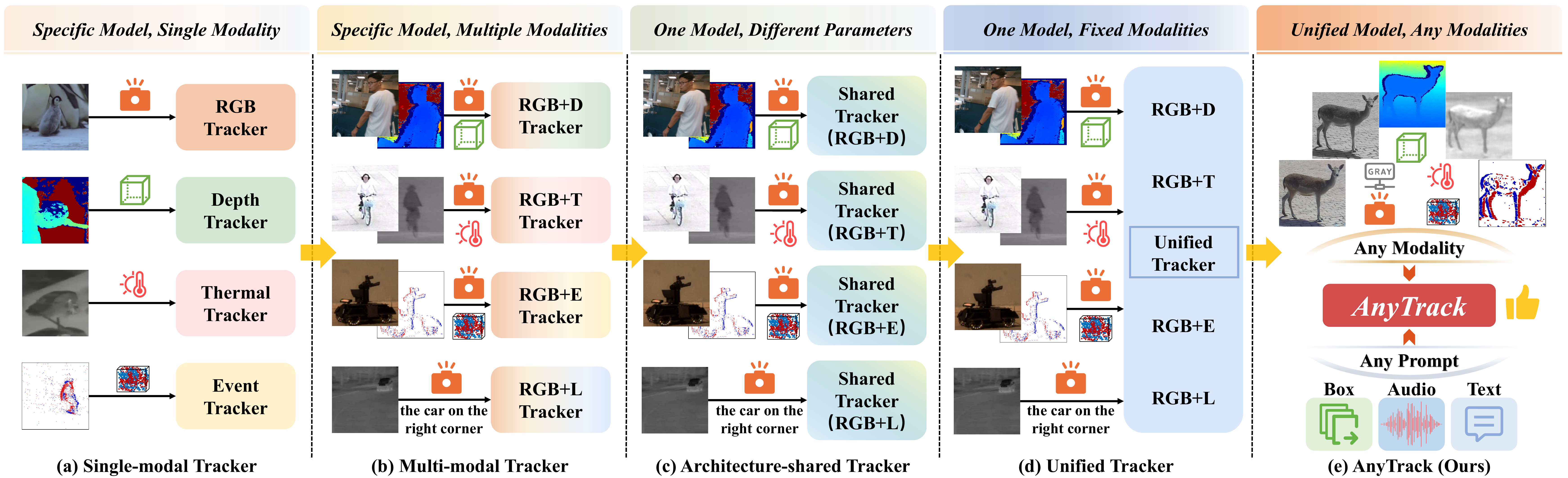}
  \vspace{-5mm}
  \caption{Comparison with different object tracking paradigms. (a) Single-modal tracker uses a separate model for each individual modality. (b) Multi-modal tracker employs specific models for fixed modality combinations. (c) Architecture-shared tracker uses one model with task-specific parameters. (d) Unified tracker supports a fixed set of modalities. (e) Our framework enables object tracking with any modalities through a unified model and flexible prompts.}
  \label{fig:motivation}
\end{teaserfigure}


\maketitle

\section{Introduction}

Visual object tracking is a fundamental task in computer vision~\cite{du2026unsupervised,du2026pansharpening,zhong2025semi,zhong2025adaptive,feng2026trimrg,FENG2026104289,shi2026culture,huang2026exposurebiasalleviatedirectional,zhao2026resilphase,zhang2026gpr,yang2026entropy} that aims to locate objects of interest across consecutive video frames.
It has important applications in autonomous driving~\cite{lan2022instance,xu2026psgait,wu2025novel,yu2026depth}, surveillance~\cite{li2026combiner,ConeSep,long2025revisiting,long2026towards,liu2025dcevo}, and robotics~\cite{wu2026promsa,xing2026adaptsplat,xia2024vit}.
Early studies focused on single-modal object tracking using individual information sources such as visible (RGB), grayscale (G), depth (D), thermal (T) infrared, or event (E) streams.
Each modality captures specific visual cues, but relying on a single source remains insufficient for robust object tracking across diverse scenarios.
This limitation has driven the development of multi-modal object tracking, which combines complementary modalities.
The field now covers various modality combinations such as RGB+D, RGB+T, RGB+E, and RGB+Language (L).
By integrating multiple information sources, multi-modal object tracking achieves more robust performance in complex conditions.

Despite great progress, existing multi-modal object tracking methods are typically designed for specific modality combinations, as shown in Fig.~\ref{fig:motivation}.
They suffer from several significant limitations.
(1)~Each combination requires a separately trained model with a customized architecture, leading to redundant parameters, inefficient resource utilizations, and complex training pipelines.
(2)~Multi-modal data streams are often imperfect due to sensor failures, synchronization issues, or temporary dropouts.
Current task-specific models lack the robustness when handling missing modality or quality degradation.
(3)~Training on isolated datasets for single tasks prevents the model from benefiting from other datasets, which constrains the generalization capability and amplifies overfitting risks.
(4)~Real-world applications demand the flexibility to handle any modalities across scenarios.
Although initial attempts have been made to unify certain tracking tasks, their scope remains limited.
For example, some approaches~\cite{vipt,chan2025smstracker,hu2025adaptive} unify the architectural design, while others~\cite{zhang2025tracking,zhang2025sam,chen2025sutrack} support a predefined set of input modalities.
Notably, these methods require RGB as an essential input and fail to operate without it.
Consequently, there is still no tracker capable of handling any combinations of diverse modalities within a single model.
This limitation raises a critical question: \textbf{\textit{Can we develop a unified framework that performs robust object tracking with flexible inputs?}}

However, constructing such a unified object tracking framework inevitably faces several challenges.
First, achieving robust tracking under any modalities is fundamentally difficult, especially when certain modalities are missing or corrupted.
Second, maintaining temporal coherence among frames across different modalities remains challenging due to the scale variation, motion blur, and similar objects.
Third, we observe that blindly fusing more modalities often degrades tracking accuracy, failing to leverage complementary cues.
Consequently, achieving heterogeneous feature interaction is non-trivial.
Finally, existing multi-modal object tracking datasets are primarily annotated with bounding boxes, lacking diverse annotations such as language descriptions or audio (A) clips.
It poses a substantial barrier to training a comprehensively unified model.

\begin{figure*}[t]
\centering
\includegraphics[width=0.99\linewidth]{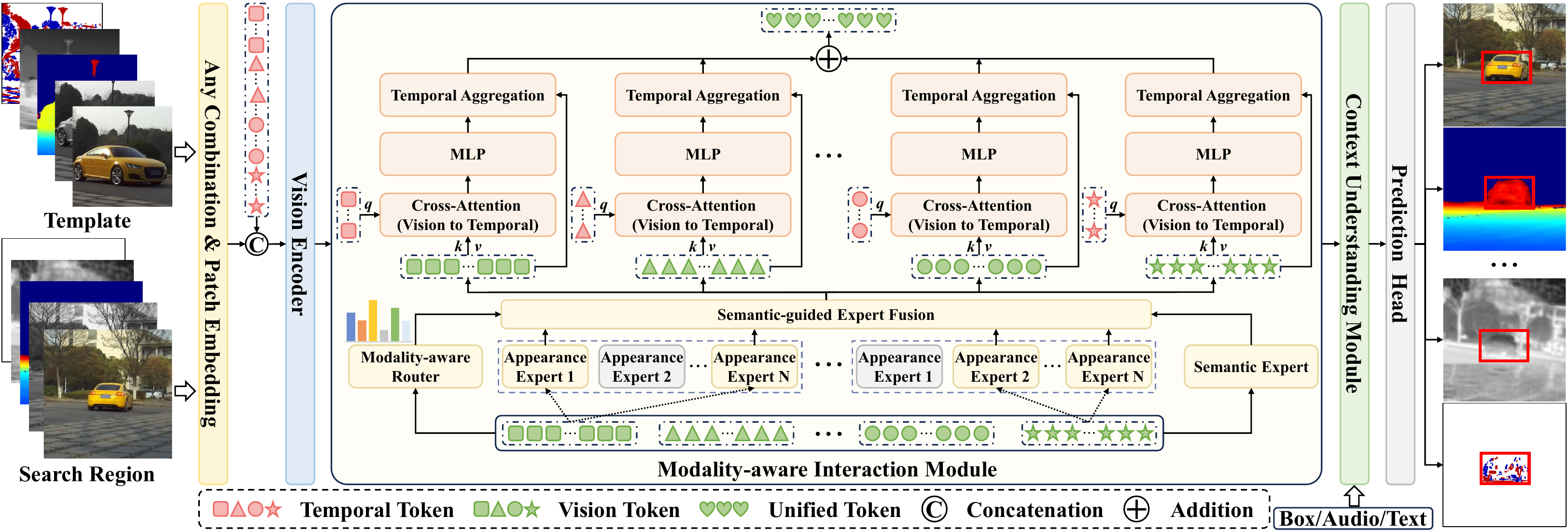}
\vspace{-2mm}
\caption{Overall framework. Firstly, template and search region images of any modalities are tokenized to form vision tokens, which are then concatenated with temporal tokens from previous frames. Then, these tokens are processed by the modality-shared vision encoder for feature extraction. Subsequently, MIM performs dynamic feature interaction while aggregating temporal information to ensure spatio-temporal consistency. Afterwards, CUM constructs global-local prompts from multi-modal references to enable target-aware context modeling. Finally, a prediction head is used for target localization.}
\label{fig:pipline}
\vspace{-3mm}
\end{figure*}

To address the aforementioned challenges, we propose \textbf{AnyTrack}, a unified and flexible framework for object tracking with any modalities.
As illustrated in Fig.~\ref{fig:pipline}, AnyTrack incorporates two core modules: \textbf{Modality-aware Interaction Module (MIM)} and \textbf{Context Understanding Module (CUM)}.
MIM facilitates dynamic interaction among diverse modalities to bridge modality discrepancies.
Through cross-modal temporal aggregation, it maintains spatio-temporal consistency within heterogeneous fusion.
Meanwhile, CUM associates visual features with target locations using global-local prompts.
Leveraging asymmetric bidirectional attention, it employs target-aware context modeling to enhance foreground-background discrimination.
To support the training and evaluation, we extend existing multi-modal tracking benchmarks with grayscale images as well as language and audio annotations.
Extensive experiments demonstrate that AnyTrack achieves superior performance in both complete and missing modality scenarios.

In summary, our main contributions are as follows:
\begin{itemize}
\item To the best of our knowledge, our AnyTrack is the first unified tracking framework that handles any combinations of diverse modalities within a single model.
\item We design a Modality-aware Interaction Module (MIM) to perform dynamic interaction, bridging modality discrepancies and maintaining spatio-temporal consistency.
\item We introduce a Context Understanding Module (CUM) that leverages global-local prompts to conduct target-aware context modeling, enhancing target discrimination.
\item We extend existing multi-modal object tracking benchmarks to enable comprehensive training and evaluation. Extensive experiments validate the effectiveness and robustness of AnyTrack across diverse tracking scenarios.
\end{itemize}

\section{Related Works}
\subsection{Single-Modal Object Tracking}
Single-modal object tracking localizes targets using visual cues from a single sensor, typically formulated as a template-to-candidate matching problem.
RGB tracking represents the dominant paradigm.
Early works~\cite{henriques2014high,bertinetto2016fully} utilized correlation filters or Siamese networks to measure similarities between the target template and search regions.
Recent methods~\cite{ding2025target,ostrack,chen2023seqtrack,cui2024mixformer,kang2025exploring,zheng2022leveraging,zheng2025decoupled,zheng2024odtrack,zheng2026boosting} employ Transformer architectures to jointly extract and fuse features for  target localization.
Beyond RGB, alternative modalities provide complementary information to address specific challenges.
Grayscale tracking~\cite{lv2018mean} remains valuable in specialized domains such as military applications and low-light environments.
Depth tracking~\cite{yan2021depth} utilizes depth maps to capture spatial occlusion cues and achieve inherent robustness against illumination variations.
Thermal infrared tracking~\cite{xie2025cst} excels under camouflage and poor illumination.
Event tracking~\cite{wang2025mambaevt} has recently emerged with bio-inspired cameras, evolving from frame conversion methods to end-to-end frameworks.
However, current single-modal trackers rely on separate models, hindering knowledge transfer across modalities.
In contrast, our method unifies diverse tracking tasks under one model through global-local prompts, enabling comprehensive target-aware context modeling without task-specific designs.

\subsection{Multi-Modal Object Tracking}
Multi-modal object tracking has gained widespread attention by leveraging complementary information across diverse modalities.
Early methods primarily use auxiliary modalities to mitigate limitations of RGB-only tracking in challenging scenarios.
These methods focus on effective feature fusion for bi-modal combinations such as RGB+D~\cite{ying2026amtrack,yan2021depthtrack}, RGB+T~\cite{li2025cadtrack,ding2026adaptive,hu2026curriculum}, RGB+E~\cite{sun2025exploring,sun2026aligntrack}, and RGB+L~\cite{zheng2026learning,zheng2023toward}.
Despite these advances, bi-modal trackers still struggle in real-world scenarios.
Recent efforts have expanded to tri-modal and even quad-modal ones.
For instance, RDTTrack~\cite{zhu2025collaborating} introduces a prompt-learning framework for RGB+D+T tracking.
SPL~\cite{zhu2024unimod1k} handles RGB+D, RGB+L, and RGB+D+L tracking, while WebUAV-3M~\cite{zhang2022webuav} further enables RGB+L+A tracking in UAV scenarios.
Additionally, RAGTrack~\cite{li2026ragtrack} introduces a retrieval-augmented generation framework for RGB+T+L tracking.
More recently, QuadFusion~\cite{lu2025towards} achieves RGB+T+E+L tracking through a multi-scale fusion mechanism.
However, existing multi-modal trackers struggle to resolve modality discrepancies and maintain temporal coherence across frames.
In contrast, our method enables modality-aware interaction among fine-grained features and integrates temporal cues to ensure robust spatio-temporal consistency.

\subsection{Unified Object Tracking}
Recent advances in foundation models~\cite{syed2026llm,shi2026tracerouter,zhang2025knowledge,xu2025lgc,xiao2026staying,xiao2026layer,huang2026detecting,shan20263d}, and the growing demand for multi-task solutions have motivated the development of unified object tracking frameworks.
These methods attempt to unify diverse object tracking tasks under a shared architecture.
ProTrack~\cite{yang2022prompting} pioneers prompt-tuning with an architecture-shared design for RGB-X tracking.
Subsequent works~\cite{hu2025exploiting,vipt,SDSTrack,chan2025smstracker} also adopt shared architectures but retrain separate parameters for each task.
To achieve a more comprehensive unification, recent trackers~\cite{tan2024xtrack,tan2025you,chen2025sutrack,Un-Track,Zheng2025umodtrack,ma2025unisot,zhang2025sam} learn a common latent space for different modality inputs, enabling parameter sharing across diverse tracking scenarios.
However, these unified trackers remain constrained to predefined modality combinations.
In contrast, our method provides the flexibility to handle any modalities within a single model, enabling effective cross-modal interaction without any prior knowledge of modality combinations.

\section{Methods}

In this paper, we propose AnyTrack for object tracking with any modalities, which includes two key modules: Modality-aware Interaction Module (MIM) and Context Understanding Module (CUM).
The overall framework is shown in Fig.~\ref{fig:pipline}.
MIM facilitates dynamic interaction across diverse modalities, bridging modality discrepancies and aggregating temporal cues to maintain spatio-temporal consistency.
CUM establishes spatial correspondence between visual features and target locations via global-local prompts, employing target-aware context modeling to enhance foreground-background discrimination.
Details are described as follows.

\subsection{Overall Framework}

The object tracking task is formulated as continuously estimating the state of objects of interest in subsequent video frames given initial templates.
At time step $t$, we take template images $\mathbf{Z}_m^t \in \mathbb{R}^{C_{Z} \times H_{Z} \times W_{Z}}$ and search regions $\mathbf{S}_m^t \in \mathbb{R}^{C_{S} \times H_{S} \times W_{S}}$ as inputs for any modality $m$, such as RGB, G, D, T, and E.
First, we transform these images into sequences of patch tokens $\hat{\mathbf{Z}}_m^t \in \mathbb{R}^{N_{Z} \times C}$ and $\hat{\mathbf{S}}_m^t \in \mathbb{R}^{N_{S} \times C}$ via three-stage downsampling~\cite{zhang2023hivit}, where $N_{Z}$ and $N_{S}$ denote the number of tokens, and $C$ is the channel dimension.
Then, we concatenate these tokens along the token dimension to obtain the tokenized representation $\mathbf{X}_m^t = [\hat{\mathbf{Z}}_m^t; \hat{\mathbf{S}}_m^t]$.
Next, we construct modality-specific representations by incorporating temporal information.
Specifically, we concatenate $\mathbf{X}_m^t$ with temporal tokens $\mathbf{H}_m^t \in \mathbb{R}^{N_{H} \times C}$ aggregated from previous frames to form $\hat{\mathbf{X}}_m^t = [\mathbf{H}_m^t; \mathbf{X}_m^t]$.
Subsequently, these tokens pass through the modality-shared vision encoder $\mathbf{E}_V$ for feature extraction:
\begin{align}
\mathbf{V}_m^t = \mathbf{E}_V(\hat{\mathbf{X}}_m^t).
\end{align}

The encoded features are then processed by MIM to enable dynamic feature interaction.
Afterwards, CUM constructs global-local prompts from multi-modal references such as motion trajectories $\mathbf{B}^{t-1}$, language descriptions $\mathbf{L}$, and audio clips $\mathbf{A}$.
With these prompts, CUM performs target-aware context modeling to enhance the feature representation.
Finally, we extract the search region features from the enhanced representation and feed them into the prediction head to obtain tracking results ${\mathbf{B}}^t$.

\subsection{Modality-aware Interaction Module}

Existing modality interaction methods face two critical limitations.
First, they rely on fixed architectures~\cite{hu2025exploiting,chen2025sutrack} for specific modality combinations, preventing flexible adaptation to any modalities.
Second, single static architectures~\cite{tbsi,Un-Track} cannot effectively handle the diverse visual characteristics of heterogeneous modalities while maintaining temporal coherence across frames.
Therefore, we propose the MIM to bridge modality discrepancies and facilitate dynamic interaction based on the Mixture-of-Experts (MoE)~\cite{shazeer2017outrageously}.
Specifically, our MIM aggregates temporal cues through cross-modal attention to ensure robust spatio-temporal consistency.

As illustrated in Fig.~\ref{fig:pipline}, our proposed MIM first performs channel-wise average pooling $\mathcal{P}$ on the encoded features $\mathbf{V}_m^t$ to obtain compact representations.
Then, they are fed into the modality-aware router $\mathcal{R}$ to obtain routing logits $\mathbf{z}_m$ across all experts:
\begin{align}
\mathbf{z}_m = {\rm{{\mathcal R}}}(\mathcal{P}(\mathbf{V}_m^t)),
\end{align}
where ${\rm{{\mathcal R}}}$ is an Multilayer Perceptron (MLP).
To ensure a diverse expert utilization during training, we employ noisy gating. Specifically, we inject learnable noise into the routing logits:
\begin{align}
\tilde{\mathbf{z}}_m &= \mathbf{z}_m + \boldsymbol{\xi} \odot (\text{Softplus}({\rm{{\mathcal N}}}(\mathcal{P}(\mathbf{V}_m^t))) + \epsilon),
\end{align}
where $\mathcal{N}$ shares the same architecture as $\mathcal{R}$, $\epsilon$ is a small constant ensuring numerical stability, $\boldsymbol{\xi} \sim \mathcal{N}(0, \mathbf{I})$ represents the standard Gaussian random noise, and $\odot$ denotes element-wise multiplication.

We then select the top-$k$ experts with highest weights for each modality and apply softmax to obtain the gating weights $\mathbf{g}_m$:
\begin{align}
\mathbf{g}_m^{(i)}= \text{Softmax}(\text{TopK}(\tilde{\mathbf{z}}_m,k))_i,\quad i \in \mathcal{K}_m,
\end{align}
where $\mathcal{K}_m$ denotes the index set of the selected experts. Crucially, all tokens of each modality are only processed by these selected experts, realizing sparse adaptive selection that adapts to modality characteristics without activating the full expert pool.

To capture both unique and shared patterns, we separate the selected experts into two complementary branches: explicit appearance expert and implicit semantic expert.
The explicit appearance experts $\mathcal{E}_i$ capture modality-specific appearance characteristics.
These characteristics emerge naturally from data-driven learning rather than manual definition, decoupling heterogeneous modality features into fine-grained appearance subspaces:
\begin{align}
\mathbf{h}_m^{(i)}=  \mathcal{E}_i(\mathbf{V}_m^t),
\end{align}
where $\mathcal{E}_i$ represents the $i$-th expert implemented as an MLP.

Meanwhile, we observe that while object appearances vary significantly across modalities, object semantics remain modality-invariant.
To exploit this invariant property and capture cross-modal shared patterns, all modalities share one implicit semantic expert $\mathcal{E}_s$ that aligns and enhances modality-invariant semantic cues.
It employs a gating mechanism to emphasize salient semantics while suppressing irrelevant backgrounds:
\begin{gather}
\tilde{\mathbf{h}}_m =\mathcal{E}_s(\mathbf{V}_m^t),\\
\mathcal{E}_s(x)=\mathcal{M}(\mathcal{C}(\mathcal{M}(\mathcal{O}(x)))\odot\delta(\mathcal{M}(\mathcal{O}(x)))),
\end{gather}
where $\mathcal{O}$ is the layer normalization~\cite{ba2016layer}, $\mathcal{M}$ is an MLP, $\mathcal{C}$ is the convolution operation, and $\delta$ is the GELU activation~\cite{hendrycks2016gaussian}.

Afterwards, we perform semantic-guided expert fusion $\mathcal{F}$ to integrate fine-grained appearance features from the activated explicit experts and cross-modal semantic features from the shared implicit expert.
To effectively combine these complementary features,
it dynamically balances the contributions of selected experts while preserving the original feature flow through residual connections, yielding the modality-specific and enhanced visual features $\mathbf{G}_m^t$:
\begin{align}
\mathbf{G}_m^t &= \mathcal{F}\left(\mathbf{V}_m^t, \sum_{i \in \mathcal{K}_m} \mathbf{g}_m^{(i)} \cdot \mathbf{h}_m^{(i)} + \tilde{\mathbf{h}}_m\right), \\
\mathcal{F}(x_1,x_2) &=x_1+\mathcal{M}(\mathcal{M}(\mathcal{O}(x_1))\odot\delta(\mathcal{M}(\mathcal{O}(x_2)))).
\end{align}

To maintain temporal coherence across modalities, we propagate the information from enhanced visual features to temporal tokens via a multi-head cross-attention $\Phi$~\cite{vit}:
\begin{align}
\tilde{\mathbf{H}}_m^t = \mathcal{M}(\Phi(\mathbf{H}_m^t, \mathbf{G}_m^t, \mathbf{G}_m^t)).
\end{align}

Finally, we perform cross-modal temporal aggregation to integrate both spatial representations and historical temporal context.
It weights enhanced visual features $\mathbf{G}_m^t$ by the affinity with refined temporal tokens $\tilde{\mathbf{H}}_m^t$ across all modalities:
\begin{align}
\mathbf{U}^t = \sum_{m} \mathbf{G}_m^t \odot (\mathbf{G}_m^t \otimes (\tilde{\mathbf{H}}_m^t)^\top),
\end{align}
where $\mathbf{U}^t$ represents the unified features, $\otimes$ denotes matrix multiplication, and $\tilde{\mathbf{H}}_m^t$ are utilized for the next frame.

Our MIM employs dynamic routing to adaptively process diverse modalities, decomposing heterogeneous features through explicit appearance experts while extracting shared semantic features via the implicit semantic expert.
This design enables fine-grained representation learning alongside cross-modal alignment through invariant semantic cues.
With cross-modal temporal aggregation, this module ensures robust spatio-temporal consistency.

\begin{figure}[t]
\centering
\includegraphics[width=0.99\linewidth]{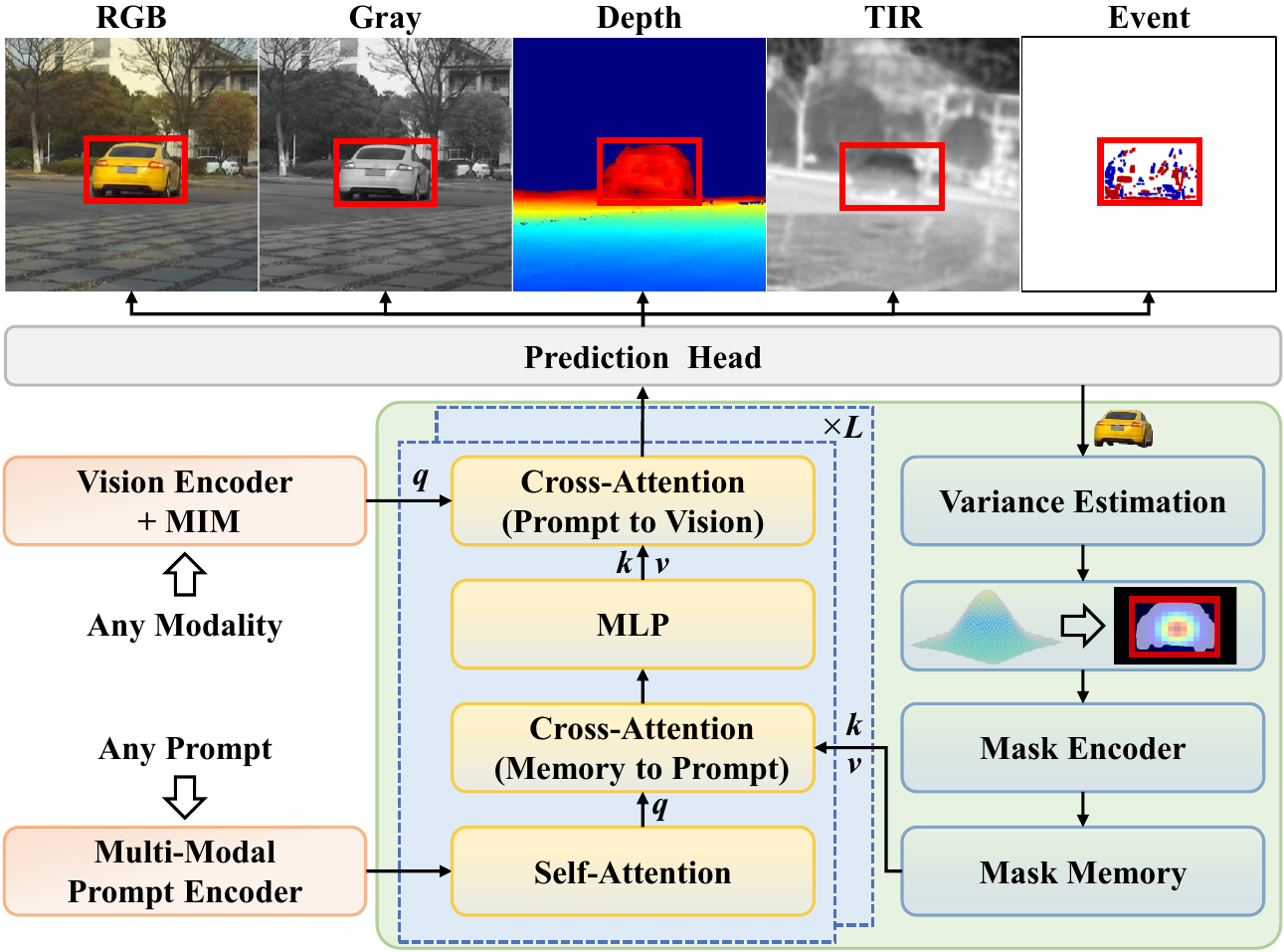}
\vspace{-2mm}
\caption{Details of our proposed CUM.}
\label{fig:AIUM}
\vspace{-4mm}
\end{figure}

\subsection{Context Understanding Module}

To precisely localize targets, it is essential to leverage multi-modal references to provide spatial priors and semantic guidance.
However, existing methods~\cite{zheng2026learning,zheng2023toward} typically integrate reference information through simple concatenation or shallow fusion, struggling to establish precise spatial correspondence and often resulting in ambiguous target boundaries.
Therefore, we propose the CUM, which establishes spatial correspondence between visual features and target locations via global-local prompts to enable target-aware context modeling.
As illustrated in Fig.~\ref{fig:AIUM}, our CUM is implemented through three key designs: multi-modal prompt encoder, asymmetric bidirectional attention, and dynamic mask generation.

\textbf{Multi-modal prompt encoder.}
First, we employ modality-specific encoders to extract features from motion trajectories $\mathbf{B}^{t-1}$, language descriptions $\mathbf{L}$, and audio clips $\mathbf{A}$:
\begin{align}
\tilde{\mathbf{B}}^t, \tilde{\mathbf{L}}, \tilde{\mathbf{A}} = \mathcal{M}(\mathbf{E}_P(\mathbf{B}^{t-1}, \mathbf{L}, \mathbf{A})),
\end{align}
where $\mathbf{E}_P$ denotes motion, text, and audio encoders, respectively. $\mathcal{M}$ is an MLP for dimension alignment.
To distinguish corner point types in motion tokens, we introduce learnable point embeddings~\cite{zhao2025efficient}.
We then concatenate these tokens along the token dimension to form the global-local prompts $\mathbf{P}^t \in \mathbb{R}^{(N_{B}+N_{L}+N_{A}) \times C}$:
\begin{align}
\mathbf{P}^t = [\tilde{\mathbf{B}}^t; \tilde{\mathbf{L}}; \tilde{\mathbf{A}}].
\end{align}

\textbf{Asymmetric bidirectional attention.}
To fully exploit cross-modal information, we model distinct information flows between global-local prompts and visual features through asymmetric bidirectional attention.
Unlike traditional unidirectional cross-attention, our mechanism enables visual features to perceive semantic guidance from prompts while allowing prompts to adapt to spatial details in visual features.
Furthermore, we preserve spatial coherence by maintaining a mask memory $\mathcal{Q}$ storing encoded target representations from previous frames.
The global-local prompts serve as a bridge between the historical spatial context and current visual features, providing essential priors for precise localization.

Initially, the global-local prompts undergo self-attention to explore intrinsic relationships.
Afterwards, we establish the \textit{memory-to-prompt} propagation, where global-local prompts $\mathbf{P}^t$ serve as queries and the mask memory $\mathcal{Q}$ provides keys and values:
\begin{align}
\tilde{\mathbf{P}}^t = \mathcal{M}(\Phi(\Gamma(\mathbf{P}^t,\mathbf{P}^t,\mathbf{P}^t), \mathcal{Q}, \mathcal{Q})),
\end{align}
where $\Gamma$ denotes the multi-head self-attention~\cite{vit}.
Subsequently, we perform the \textit{prompt-to-vision} propagation, using unified features $\mathbf{U}^t$ as queries and prompt features $\tilde{\mathbf{P}}^t$ as keys and values:
\begin{align}
\tilde{\mathbf{U}}^t = \Phi(\mathbf{U}^t, \tilde{\mathbf{P}}^t, \tilde{\mathbf{P}}^t).
\end{align}

This asymmetric bidirectional attention iterates for $L$ layers.
For each cross-attention, we incorporate dense spatial maps~\cite{tancik2020fourier} as the positional encoding to visual features and the mask memory, ensuring spatial correspondence between modalities.

\textbf{Dynamic mask generation.}
To adapt to sequence-specific target variations, we further generate position-adaptive Gaussian pseudo-masks to provide flexible spatial guidance for target localization.
Given the tracking bounding box $\mathbf{B}^t$ with center $(x_c^t, y_c^t)$, width $w^t$, and height $h^t$, we initialize the Gaussian variances as $\tilde{\sigma}_{x} = w^t/6$ and $\tilde{\sigma}_{y} = h^t/6$.
While this static initialization provides a reasonable spatial prior, we dynamically refine them to accommodate sequence-specific target variations.
Specifically, a variance estimation network $\mathcal{V}$ predicts the offsets based on $\tilde{\mathbf{U}}^t$:
\begin{align}
\Delta\sigma_{x}, \Delta\sigma_{y} = \mathcal{V}(\mathcal{P}(\tilde{\mathbf{U}}^t)).
\end{align}

The dynamic variances are computed as $\sigma = \tilde{\sigma} + \Delta\sigma$, yielding position-adaptive Gaussian pseudo-masks:
\begin{align}
\mathbf{D}^t(x, y) \propto \exp\left(-\frac{(x-x_c^t)^2}{2\sigma_x^2} - \frac{(y-y_c^t)^2}{2\sigma_y^2}\right).
\end{align}

Finally, the pseudo-mask $\mathbf{D}^t$ is encoded via  the mask encoder $\mathcal{I}$ and the feature fusion network $\mathcal{T}$ to yield refined mask features $\mathbf{Y}^t$, which are stored in the mask memory $\mathcal{Q}$ for the next frame:
\begin{gather}
\mathbf{Y}^t = \mathcal{T}(\tilde{\mathbf{U}}^t + \mathcal{I}(\mathbf{D}^t)),\\
\mathcal{T}(x)=\mathcal{C}(\delta(\mathcal{C}(\mathcal{O}(\mathcal{C}(x))))),\quad\mathcal{I}(x)=\mathcal{C}(\delta(\mathcal{O}(\mathcal{C}(x)))).
\end{gather}

Our CUM establishes spatial correspondence between visual features and target locations via global-local prompts.
By hierarchically propagating information from the mask memory through prompts to vision features, it achieves bidirectional semantic calibration and refines target-aware context modeling.
This module ensures that prompts effectively guide visual features to focus on target regions, enhancing foreground-background discrimination.

\subsection{Prediction Head and Loss Function}

\textbf{Prediction head.}
We extract search region features from the refined representation $\tilde{\mathbf{U}}^t$ and feed them into a prediction head.
This head produces three outputs: classification scores, spatial offsets, and normalized sizes.
The tracking result $\mathbf{B}^t$ is obtained by selecting the location with the peak classification response.

\textbf{Loss function.}
To ensure a balanced expert utilization, we incorporate a balancing loss $\mathcal{L}_{\text{balance}}$.
Specifically, we define the importance score $\mathbf{I}_m^{(i)}$ for expert $\mathcal{E}_i$ as the sum of gating weights across batch samples indexed by $b$:
\begin{align}
\mathbf{I}_m^{(i)} = \sum_{b} \mathbf{g}_m^{(b,i)}.
\end{align}

We define the load $\mathbf{d}_m^{(i)}$ as the activation frequency of expert $\mathcal{E}_i$:
\begin{align}
\mathbf{d}_m^{(i)} = \sum_{b} p_m^{(b,i)}.
\end{align}
Specifically, since the discrete top-$k$ operation is non-differentiable, we employ the Gaussian cumulative distribution function~\cite{shazeer2017outrageously} to compute $p_m^{(b,i)}$ from the routing logits $\tilde{\mathbf{z}}_m$.
This probability represents the likelihood that expert $\mathcal{E}_i$ falls within the top-$k$ selection.
This formulation enables the gradient flow through the discrete routing decision.
The balancing loss is formulated as the sum of importance and load terms across all modalities:
\begin{align}
\mathcal{L}_{\text{balance}} = \sum_{m}\left(\phi(\{\mathbf{I}_m^{(i)}\}_{i=1}^N) + \phi(\{\mathbf{d}_m^{(i)}\}_{i=1}^N)\right),\\
\phi(\mathbf{x}) = \frac{\mathbb{E}[(\mathbf{x}-\mathbb{E}[\mathbf{x}])^2]}{(\mathbb{E}[\mathbf{x}])^2 + \beta}, \quad
\mathbb{E}[\mathbf{x}] = \frac{1}{N}\sum_{i=1}^N x_i,
\end{align}
where $\phi$ denotes the coefficient of variation~\cite{shazeer2017outrageously}, and $\beta$ is a small constant.
This constraint promotes the balanced expert utilization across modalities, enhancing training stability and generalization.

For optimization, we adopt a multi-task loss function comprising focal loss $\mathcal{L}_{\text{cls}}$ for classification, $L_1$ loss $\mathcal{L}_1$ and generalized IoU loss $\mathcal{L}_{\text{iou}}$ for regression, and balancing loss $\mathcal{L}_{\text{balance}}$ for regularizing expert routing.
The overall loss is formulated as:
\begin{equation}
\mathcal{L} = \lambda_{\text{cls}}\mathcal{L}_{\text{cls}} + \lambda_{\text{iou}} \mathcal{L}_{\text{iou}} + \lambda_{1} \mathcal{L}_1 + \lambda_{\text{balance}} \mathcal{L}_{\text{balance}},
\end{equation}
where $\lambda_{\text{cls}}$, $\lambda_{\text{iou}}$, $\lambda_{1}$, and $\lambda_{\text{balance}}$ are the hyper-parameters.

\section{Experiments}

\subsection{Datasets and Evaluation Metrics}

We evaluate AnyTrack on four multi-modal object tracking benchmarks.
RGBDT500~\cite{zhu2025collaborating} contains 500 video sequences across 66 object categories, with 400 for training and 100 for testing.
This benchmark provides aligned RGB, depth, and thermal images for all frames, with tracking accuracy quantified via Distance Precision (DP) and Area Under Curve (AUC) of success plots.
LasHeR~\cite{li2021lasher} is a short-term RGB-T dataset with 979 training and 245 testing video pairs annotated with 19 fine-grained attributes.
Evaluation metrics include Precision Rate (PR), Normalized Precision Rate (NPR), and Success Rate (SR).
VisEvent~\cite{wang2023visevent} provides 500 training and 320 testing video pairs for RGB-Event tracking with 17 challenge attributes, where the performance is measured via PR and SR.
DepthTrack~\cite{yan2021depthtrack} is a long-term RGB-D tracking benchmark with 150 training and 50 testing sequences, each annotated with 15 challenge attributes.
Trackers are evaluated based on PR, Recall (RE), and F-score.

\subsection{Multi-modal Benchmark Extension}

To support comprehensive training and evaluation under diverse inputs, we extend existing multi-modal object tracking benchmarks with grayscale images, language descriptions, and audio clips.
Specifically, we convert all RGB images to grayscale via standard luminance weighting to simulate low-light or monochrome conditions.
Simultaneously, we annotate each video sequence with language and audio modalities describing the tracking target.
To generate high-quality annotations, we employ a two-stage pipeline utilizing Multi-modal Large Language Models (MLLM)~\cite{Qwen2VL} followed by human verification.
For language descriptions, the MLLM receives the first frame and object bounding box with a prompt requesting a concise noun phrase formatted as ``A [object category]''.
For audio clips, a detailed prompt synthesizes appearance features (including color, object type, and distinguishing features) and dynamic states (such as motion or pose) into a coherent spoken sentence.
Subsequently, human experts review the generated content to rectify hallucinations, grammatical errors, and mixed-language artifacts, ensuring faithful representations of the visual target.
Finally, we convert the refined text into audio via API-based text-to-speech synthesis~\cite{wang2026ernie}.
Through this pipeline, we produce 2,744 language descriptions, 2,744 audio clips, and 1,604.1K grayscale frames.

\begin{table}[t]
\footnotesize
\centering
\caption{Performance comparison on RGBDT500.}
\vspace{-3mm}
    \renewcommand\arraystretch{0.8}{
	\resizebox{\linewidth}{!}{
\begin{tabular}{ccccc}
\toprule
Method & Source & Input Modality & AUC $\uparrow$   & DP $\uparrow$ \\ \hline
OSTrack~\cite{ostrack} & ECCV 2022    & RGB     & 69.4  &  73.6     \\
SeqTrack~\cite{chen2023seqtrack} & CVPR 2023   & RGB  & 71.9   & 76.6   \\
MixFormer~\cite{cui2024mixformer}  & TPAMI 2024  & RGB  & 73.2   & 78.1  \\
\rowcolor[RGB]{254,236,221}
AnyTrack  & Ours  & RGB &  77.3  &  83.7 \\
\noalign{\hrule height 0.5pt}
DeT\_ATOM~\cite{yan2021depthtrack} & ICCV 2021  & RGB+D & 63.9   & 66.5  \\
DeT\_DiMP~\cite{yan2021depthtrack}& ICCV 2021  & RGB+D & 66.7   & 70.0 \\
ViPT~\cite{vipt} & CVPR 2023 & RGB+D   & 72.0   & 75.9  \\
SDSTrack~\cite{SDSTrack} & CVPR 2024 & RGB+D  & 71.8   & 76.3\\
{Un-Track}~\cite{Un-Track}& CVPR 2024 & RGB+D & 73.3   & 77.6 \\
\rowcolor[RGB]{254,236,221}
AnyTrack  & Ours  & RGB+D  &  77.8  &  84.5 \\
\noalign{\hrule height 0.5pt}
TBSI~\cite{tbsi}  & CVPR 2023   & RGB+T  & 69.0   & 74.9  \\
ViPT~\cite{vipt} & CVPR 2023   & RGB+T & 69.3   & 75.2\\
SDSTrack~\cite{SDSTrack}  & CVPR 2024  & RGB+T    & 66.6   & 70.8 \\
BAT~\cite{BAT2024}  & AAAI 2024  & RGB+T & 71.3   & 78.2 \\
{Un-Track}~\cite{Un-Track} & CVPR 2024  & RGB+T & 73.2   & 79.0                \\
DCEvo+OSTrack~\cite{liu2025dcevo} & CVPR 2025 & RGB+T  & 70.6   & 74.1  \\
\rowcolor[RGB]{254,236,221}
AnyTrack  & Ours  & RGB+T  &  78.0  & 84.7  \\
\noalign{\hrule height 0.5pt}
RDTTrack~\cite{zhu2025collaborating} & NeurIPS 2025 & RGB+D+T  & 75.2 & 79.2 \\
\rowcolor[RGB]{254,236,221}
AnyTrack  & Ours  & RGB+D+T  &  78.3  & 85.5  \\
\noalign{\hrule height 0.5pt}
\rowcolor[RGB]{254,236,221}
AnyTrack  & Ours  & D  &  14.6  &  3.3 \\
\rowcolor[RGB]{254,236,221}
AnyTrack  & Ours  & T  &  44.1  & 29.6  \\
\rowcolor[RGB]{254,236,221}
AnyTrack  & Ours  & G  &  72.9  & 80.1  \\
\noalign{\hrule height 0.5pt}
\rowcolor[RGB]{254,236,221}
AnyTrack  & Ours  & D+A  &  15.2  & 3.7  \\
\rowcolor[RGB]{254,236,221}
AnyTrack  & Ours  & T+A  &  45.6  & 31.7  \\
\rowcolor[RGB]{254,236,221}
AnyTrack  & Ours  & G+D  &  74.3  &  82.0 \\
\rowcolor[RGB]{254,236,221}
AnyTrack  & Ours  & G+T  &  74.6  & 82.6  \\
\noalign{\hrule height 0.5pt}
\rowcolor[RGB]{237,247,234}
AnyTrack  & Ours  & D+L+A  &  15.8  & 4.5  \\
\rowcolor[RGB]{237,247,234}
AnyTrack  & Ours  & T+L+A  &  46.6  &  33.2 \\
\rowcolor[RGB]{237,247,234}
AnyTrack  & Ours  & D+T+A  &  46.6  &  34.1 \\
\rowcolor[RGB]{237,247,234}
AnyTrack  & Ours  & D+T+L  &  48.2  & 37.1  \\
\rowcolor[RGB]{237,247,234}
AnyTrack  & Ours  & G+L+A  & 74.5   & 81.8  \\
\rowcolor[RGB]{237,247,234}
AnyTrack  & Ours  & G+D+T  & 75.1  & 82.8  \\
\rowcolor[RGB]{237,247,234}
AnyTrack  & Ours  & G+D+L  & 75.8   &  83.2 \\
\rowcolor[RGB]{237,247,234}
AnyTrack  & Ours  & G+T+L  & 76.0   & 83.1  \\
\rowcolor[RGB]{237,247,234}
AnyTrack  & Ours  & RGB+L+A  & 77.9   &  84.4 \\
\noalign{\hrule height 0.5pt}
\rowcolor[RGB]{254,236,221}
AnyTrack  & Ours  & G+D+T+L  &  76.1  &  83.1 \\
\rowcolor[RGB]{254,236,221}
AnyTrack  & Ours  & G+T+L+A  &  76.2  &  83.3 \\
\rowcolor[RGB]{254,236,221}
AnyTrack  & Ours  & RGB+D+T+L  &  79.0  &  85.8 \\
\noalign{\hrule height 0.5pt}
\rowcolor[RGB]{254,236,221}
AnyTrack  & Ours  & RGB+D+T+L+A  & \textbf{79.2}   &  \textbf{86.1} \\
\noalign{\hrule height 0.5pt}
\rowcolor[RGB]{254,236,221}
AnyTrack  & Ours  & RGBDT500$_{miss}$  & 67.7   &  68.5 \\
\noalign{\hrule height 0.8pt}
\end{tabular}
}
}
\label{table:rgbdt_comparison}
\vspace{-4mm}
\end{table}

\subsection{Implementation Details}

We implement AnyTrack using PyTorch and train it on 4 NVIDIA V100 GPUs.
The visual backbone employs HiViT-B initialized from Fast-ITPN~\cite{tian2024fast,li2025dynamic}.
The text encoder utilizes CLIP~\cite{radford2021learning} and the audio encoder employs WavLM~\cite{chen2022wavlm}, both kept frozen during training.
We set the feature dimension $C=512$, temporal token number $N_H=1$, motion token number $N_B=2$, language token number $N_L=1$, and audio token number $N_A=1$.
The appearance expert number $N$ is set to 6 with top-$2$ selected during routing and $\epsilon=10^{-2}$ for numerical stability.
The asymmetric bidirectional attention comprises $L=2$ layers, and the feature fusion network $\mathcal{T}$ contains 2 layers.
For audio generation, we use text-to-speech synthesis with a sampling rate of 16, 000 Hz.
Regarding training configuration, the batch size is set to 16.
Template images and search regions are resized to $128\times128$ and $256\times256$, respectively.
We employ the AdamW optimizer~\cite{AdamW} with a learning rate of $10^{-4}$ and weight decay of $10^{-4}$.
The model is trained on the combined training sets of RGBDT500, DepthTrack, LasHeR, and VisEvent with a sampling ratio of 1:1:1:1.
Standard data augmentation techniques are applied to training samples, including random rotation, translation, and color jittering.
The loss weights are set as $\lambda_{\text{cls}}=1$, $\lambda_{\text{iou}}=2$, $\lambda_{1}=5$, and $\lambda_{\text{balance}}=0.01$ with $\beta=10^{-6}$.

\begin{table}[t]
    \centering
     \caption{Performance comparison on LasHeR.}
     \vspace{-3mm}
    \renewcommand\arraystretch{0.85}{
	\resizebox{\linewidth}{!}{
    \begin{tabular}{cccccc}
    \toprule
    Method & Source & Input Modality & PR$\uparrow$  & NPR$\uparrow$  & SR$\uparrow$  \\ \hline
        ProTrack~\cite{yang2022prompting} & ACM MM 2022 & RGB+T & 50.9  & $-$  & 42.1  \\
        ViPT~\cite{vipt} & CVPR 2023 & RGB+T & 65.1 & $-$ & 52.5 \\
        TBSI~\cite{tbsi} & CVPR 2023 & RGB+T & 69.2 & 65.7 & 55.6  \\
        {Un-Track}~\cite{Un-Track} & CVPR 2024 & RGB+T & 66.7& $-$ & 53.6  \\
        SDSTrack~\cite{SDSTrack} & CVPR 2024 & RGB+T & 66.5& $-$ & 53.1 \\
        OneTracker~\cite{OneTracker} & CVPR 2024  & RGB+T & 67.2& $-$ & 53.8  \\
        BAT~\cite{BAT2024} & AAAI 2024  & RGB+T & 70.2& $-$ & 56.3  \\
    SUTrack~\cite{chen2025sutrack}  &  AAAI 2025 & RGB+T & 74.5 & $-$ & 59.9 \\
    SMSTracker~\cite{chan2025smstracker} &  ICCV 2025 & RGB+T & 70.3 & $-$ & 56.0 \\
    XTrack~\cite{tan2024xtrack}  &  ICCV 2025 & RGB+T & 69.1 & $-$ & 55.7 \\
    UM-ODTrack~\cite{Zheng2025umodtrack} &  TPAMI 2025 & RGB+T & 74.4 & $-$ & 58.8 \\
     UniSOT~\cite{ma2025unisot} &  TPAMI 2025&  RGB+T & 67.4& $-$  & 54.0 \\
    \rowcolor[RGB]{254,236,221}
AnyTrack  & Ours  & RGB+T  &  75.5  & 71.1 & 59.7 \\
\noalign{\hrule height 0.5pt}
\rowcolor[RGB]{254,236,221}
AnyTrack  & Ours  & T  &  60.0  & 53.2 & 45.1 \\
\rowcolor[RGB]{254,236,221}
AnyTrack  & Ours  & G &  63.1  & 58.2 & 49.5 \\
\rowcolor[RGB]{254,236,221}
AnyTrack  & Ours  & RGB  &  69.4  & 64.6  & 54.8 \\
\noalign{\hrule height 0.5pt}
\rowcolor[RGB]{237,247,234}
AnyTrack  & Ours  & T+L  &  61.0  & 54.9 & 46.7\\
\rowcolor[RGB]{237,247,234}
AnyTrack  & Ours  & T+A  & 61.5   & 55.0 & 46.9\\
\rowcolor[RGB]{237,247,234}
AnyTrack  & Ours  & G+A  &  64.8  & 59.8 & 50.5\\
\rowcolor[RGB]{237,247,234}
AnyTrack  & Ours  & G+L  & 67.0   & 62.2 & 51.8\\
\rowcolor[RGB]{237,247,234}
AnyTrack  & Ours  & G+T  & 73.7   & 69.6 & 58.0 \\
\noalign{\hrule height 0.5pt}
\rowcolor[RGB]{254,236,221}
AnyTrack  & Ours  & G+T+L  & 75.2   & 71.4 & 59.3\\
\rowcolor[RGB]{254,236,221}
AnyTrack  & Ours  & RGB+T+A  & 76.0  & 71.8 & 60.3 \\
\noalign{\hrule height 0.5pt}
\rowcolor[RGB]{254,236,221}
AnyTrack  & Ours  & G+T+L+A  &  75.6  & 71.7 & 59.6 \\
\rowcolor[RGB]{254,236,221}
AnyTrack  & Ours  & RGB+T+L+A  &  \textbf{77.2}  & \textbf{72.9} & \textbf{61.1} \\
\noalign{\hrule height 0.8pt}
       \end{tabular}}
       }
\label{rgbt_overall_result}
\vspace{-3mm}
\end{table}

\subsection{Comparison with State-of-the-Art Trackers}

We evaluate AnyTrack on four benchmarks against state-of-the-art methods, with detailed analysis provided below.

\begin{table}[t]
\footnotesize
\centering
\caption{Performance comparison on DepthTrack.}
\vspace{-3mm}
    \renewcommand\arraystretch{0.85}{
	\resizebox{\linewidth}{!}{
\begin{tabular}{cccccc}
\toprule
Method & Source & Input Modality & F-score~$\uparrow$ &RE~$\uparrow$ & PR~$\uparrow$  \\ \hline
ProTrack~\cite{yang2022prompting}     & ACM MM 2022    & RGB+D      & 57.8  & 57.3  & 58.3 \\
ViPT~\cite{vipt} & CVPR 2023 & RGB+D   & 59.4  & 59.6  & 59.2  \\
{Un-Track}~\cite{Un-Track} & CVPR 2024 & RGB+D & 61.2    & 61.0 & 61.3\\
SDSTrack~\cite{SDSTrack} & CVPR 2024 & RGB+D  & 61.4  & 60.9  & 61.9\\
OneTrack~\cite{OneTracker} & CVPR 2024 & RGB+D   & 60.9  & 60.4  & 60.7  \\
SUTrack~\cite{chen2025sutrack}& AAAI 2025 & RGB+D & 65.1   &
 65.7 & 64.5 \\
SMSTracker~\cite{chan2025smstracker}& ICCV 2025 & RGB+D & 63.6   & 63.1 & 64.1 \\
XTrack~\cite{tan2024xtrack}& ICCV 2025 & RGB+D & 61.5   & 62.0 & 61.8 \\
 UM-ODTrack~\cite{Zheng2025umodtrack} &  TPAMI 2025 & RGB+D & 61.2   &  62.2 & 60.3 \\
 UniSOT~\cite{ma2025unisot} &  TPAMI 2025 & RGB+D & 62.5   &  62.2 & 62.9 \\
AMTrack~\cite{ying2026amtrack}& AAAI 2026 & RGB+D & 63.4  & 62.8  & 64.1 \\
\rowcolor[RGB]{254,236,221}
AnyTrack  & Ours  & RGB+D  &  65.2  & 65.4 & 64.9 \\
\noalign{\hrule height 0.5pt}
\rowcolor[RGB]{254,236,221}
AnyTrack  & Ours  & D  & 28.0 & 25.9 & 30.4 \\
\rowcolor[RGB]{254,236,221}
AnyTrack  & Ours  & G & 50.7 & 48.7 & 52.8\\
\rowcolor[RGB]{254,236,221}
AnyTrack  & Ours  & RGB  & 62.7   & 62.1 & 63.3 \\
\noalign{\hrule height 0.5pt}
\rowcolor[RGB]{237,247,234}
AnyTrack  & Ours  & D+L  &  30.9  & 28.3 & 34.0 \\
\rowcolor[RGB]{237,247,234}
AnyTrack  & Ours  & D+A  &  29.6  & 26.8  & 33.0 \\
\rowcolor[RGB]{237,247,234}
AnyTrack  & Ours  & G+L  &  51.2  & 49.4 & 53.1 \\
\rowcolor[RGB]{237,247,234}
AnyTrack  & Ours  & G+A  & 53.2   & 50.6 & 55.9 \\
\rowcolor[RGB]{237,247,234}
AnyTrack  & Ours  & G+D  &  56.5  & 55.5  & 57.5 \\
\noalign{\hrule height 0.5pt}
\rowcolor[RGB]{254,236,221}
AnyTrack  & Ours  & G+D+A  & 56.9   & 56.1 & 57.7 \\
\rowcolor[RGB]{254,236,221}
AnyTrack  & Ours  & RGB+D+L  & \textbf{65.9}   & \textbf{65.9} & \textbf{66.0} \\
\noalign{\hrule height 0.8pt}
\end{tabular}
}
}
\label{tab-rgbd}
\vspace{-5mm}
\end{table}
\textbf{RGBDT500.}
Tab.~\ref{table:rgbdt_comparison} shows the comparative performance on RGBDT500.
AnyTrack attains 83.7\% DP with RGB input, surpassing previous RGB methods including MixFormer~\cite{cui2024mixformer}.
Meanwhile, it maintains strong performance at 72.9\% AUC with grayscale images, demonstrating robustness to color degradation.
For bi-modal tracking, AnyTrack achieves 77.8\% AUC on RGB+D and 78.0\% AUC on RGB+T, exceeding unified approaches like Un-Track~\cite{Un-Track} and specialized trackers such as TBSI~\cite{tbsi}.
With all modalities, AnyTrack delivers the best performance of 79.2\% AUC and 86.1\% DP, outperforming RDTTrack~\cite{zhu2025collaborating}.
This validates the effectiveness in fusing complementary information.
Notably, combining imperfect single-modal inputs (e.g., G+D or G+T) substantially improves performance, confirming that our framework effectively leverages cross-modal interaction even with degraded modalities.

\textbf{LasHeR.}
Tab.~\ref{rgbt_overall_result} shows the comparison results on LasHeR.
With RGB+T inputs augmented by language and audio, AnyTrack achieves 77.2\% PR, outperforming unified trackers such as SMSTracker~\cite{chan2025smstracker} and XTrack~\cite{tan2024xtrack}.
Furthermore, combining G+T with language and audio yields 59.6\% SR, surpassing specialized RGB-T methods including TBSI~\cite{tbsi} and BAT~\cite{BAT2024}.
Under single-modal settings, the method maintains competitive performance, with RGB-only tracking reaching 69.4\% PR, grayscale-only tracking attaining 63.1\% PR, and thermal-only tracking achieving 60.0\% PR, exceeding ProTrack~\cite{yang2022prompting}.
Moreover, the consistent gains across diverse modality configurations validate our framework's ability to handle missing or degraded sensors without task-specific retraining.

\textbf{DepthTrack.}
Tab.~\ref{tab-rgbd} shows the quantitative results on DepthTrack.
With RGB+D augmented by language, AnyTrack records 65.9\% F-score, 65.9\% RE and 66.0\% PR, obtaining a 2.3\% gain over SMSTracker~\cite{chan2025smstracker}.
The efficacy of multi-modal fusion is particularly pronounced when comparing combined inputs against the single-depth configuration.
Specifically, integrating grayscale with depth improves F-score by 28.5\% relative to depth-only tracking, while RGB+D further boosts performance by 37.2\%.
RGB-only tracking yields 62.7\% F-score, marking a 3.3\% improvement over ViPT~\cite{vipt} and validating strong single-modality robustness.

\begin{table}[t]
\footnotesize
\centering
\caption{Performance comparison on VisEvent.}
\vspace{-3mm}
    \renewcommand\arraystretch{0.85}{
	\resizebox{\linewidth}{!}{
\begin{tabular}{ccccc}
\toprule
Method & Source & Input Modality & SR $\uparrow$    & PR  $\uparrow$  \\ \hline
ProTrack~\cite{yang2022prompting}     & ACM MM 2022    & RGB+E      & 47.4   & 61.7 \\
ViPT~\cite{vipt} & CVPR 2023 & RGB+E   & 59.2   & 75.8  \\
{Un-Track}~\cite{Un-Track}& CVPR 2024 & RGB+E & 59.7   & 76.3 \\
SDSTrack~\cite{SDSTrack} & CVPR 2024 & RGB+E  & 59.7   & 76.7\\
OneTrack~\cite{OneTracker} & CVPR 2024 & RGB+E   & 60.8   & 76.7  \\
SUTrack~\cite{chen2025sutrack}& AAAI 2025 & RGB+E & 62.7   & 79.9 \\
MamTrack~\cite{sun2025exploring}& CVPR 2025 & RGB+E & 61.6   & 79.2 \\
SMSTracker~\cite{chan2025smstracker}& ICCV 2025 & RGB+E & 60.4   & 76.3 \\
XTrack~\cite{tan2024xtrack}& ICCV 2025 & RGB+E & 60.9   & 77.5 \\
UM-ODTrack~\cite{Zheng2025umodtrack} &  TPAMI 2025 & RGB+E & 59.6   &  78.9 \\
UniSOT~\cite{ma2025unisot} &  TPAMI 2025 & RGB+E & 60.7   &  78.0 \\
AlignTrack~\cite{sun2026aligntrack} &  AAAI 2026 & RGB+E & 63.7   & 80.4 \\
\rowcolor[RGB]{254,236,221}
AnyTrack  & Ours  & RGB+E  &  63.2  & 81.0  \\
\noalign{\hrule height 0.5pt}
\rowcolor[RGB]{254,236,221}
AnyTrack  & Ours  & E  &  29.4  &  46.3 \\
\rowcolor[RGB]{254,236,221}
AnyTrack  & Ours  & G &  57.2 & 76.8 \\
\rowcolor[RGB]{254,236,221}
AnyTrack  & Ours  & RGB  &  62.4  & 79.8  \\
\noalign{\hrule height 0.5pt}
\rowcolor[RGB]{237,247,234}
AnyTrack  & Ours  & E+L  &  30.8  &  47.3 \\
\rowcolor[RGB]{237,247,234}
AnyTrack  & Ours  & G+A  &  58.1  &  77.1 \\
\rowcolor[RGB]{237,247,234}
AnyTrack  & Ours  & G+E  & 59.2   &  78.6 \\
\noalign{\hrule height 0.5pt}
\rowcolor[RGB]{254,236,221}
AnyTrack  & Ours  & G+E+A  &  60.7  & 79.5  \\
\rowcolor[RGB]{254,236,221}
AnyTrack  & Ours  & G+E+L  & 61.2   & 79.7  \\
\rowcolor[RGB]{254,236,221}
AnyTrack  & Ours  & RGB+E+L  &  \textbf{63.9}  & \textbf{81.3}  \\
\noalign{\hrule height 0.8pt}
\end{tabular}
}
}
\label{tab-sota-rgbe}
\vspace{-5mm}
\end{table}

\begin{table}[t]
\caption{Comparison on modality-missing datasets.}
\vspace{-3mm}
\label{tab-sota-miss}
\centering
\renewcommand\arraystretch{0.85}{
	\resizebox{\linewidth}{!}{
\begin{tabular}{c|ccc|ccc|cc}
\toprule
\multirow{2}{*}{Method}
& \multicolumn{3}{c|}{LasHeR$_{miss}$}
& \multicolumn{3}{c|}{DepthTrack$_{miss}$}
& \multicolumn{2}{c}{VisEvent$_{miss}$}  \\
\cline{2-4} \cline{5-7} \cline{8-9}
& \rule{0pt}{2.25ex}PR $\uparrow$ & \rule{0pt}{2.25ex}NPR $\uparrow$ & \rule{0pt}{2.25ex}SR $\uparrow$ & \rule{0pt}{2.25ex}F-score $\uparrow$ & \rule{0pt}{2.25ex}RE $\uparrow$ & \rule{0pt}{2.25ex}PR $\uparrow$
& \rule{0pt}{2.25ex}SR $\uparrow$ & \rule{0pt}{2.25ex}PR $\uparrow$ \\
\midrule[0.5pt]
MCITrack~\cite{kang2025exploring} & 40.0& 36.5 & 32.2& 49.7 & 42.9 & 59.1  & 36.5 & 49.9  \\
ViPT~\cite{vipt} & 40.1 & 37.5 & 34.0& 44.4 & 40.5 & 46.6  & 43.2 & 57.2  \\
SeqTrackv2~\cite{chen2023seqtrack} & 50.0 & 46.2 &39.9& 45.0 & 40.9 & 50.0  & 43.1 & 57.6  \\
IPT~\cite{lu2025modality} & 61.7 & 56.8 & 49.4 & - & - & - & - & -  \\
SDSTrack~\cite{SDSTrack} & 52.5 & 48.6 &43.1& 46.7 & 42.0 & 52.7  & 46.9 & 62.6  \\
STTrack~\cite{hu2025exploiting} & 54.5 & 51.2 &44.9& 49.9 & 48.8 & 51.0  & 49.7 &  65.5 \\
SUTrack~\cite{chen2025sutrack} & 58.3 & 53.8 &47.6& 49.5 & 47.3 & 51.9  &50.5  & 66.6  \\
\noalign{\hrule height 0.5pt}
\rowcolor[RGB]{254,236,221}
AnyTrack (Ours) & \textbf{70.6} &\textbf{64.8} & \textbf{55.3} & \textbf{54.9} & \textbf{52.9} & \textbf{57.1} & \textbf{54.0} &\textbf{72.9} \\
\noalign{\hrule height 0.8pt}
\end{tabular}
}
}
\vspace{-5mm}
\end{table}

\textbf{VisEvent.}
Tab.~\ref{tab-sota-rgbe} shows the results on VisEvent.
With RGB+E augmented by language, AnyTrack achieves 81.3\% PR, outperforming event-based trackers such as MamTrack~\cite{sun2025exploring}, as well as unified frameworks including XTrack~\cite{tan2024xtrack}.
Additionally,combining G+E with language attains 61.2\% SR, demonstrating that the framework effectively captures motion boundaries from asynchronous event streams without requiring color information.
Moreover, the consistent gains observed when incorporating language and audio references validate our framework's ability to exploit auxiliary prompts for robust tracking even with limited visual sensors.

\textbf{Modality-missing datasets.}
To evaluate robustness under various modality-missing scenarios, we construct RGBDT500$_{miss}$ following established protocols~\cite{lu2025modality,tan2025you}.
This benchmark comprises three typical missing patterns to simulate real-world challenges: random missing, switched missing, and long-time missing.
Meanwhile, we evaluate on existing modality-missing benchmarks including LasHeR$_{miss}$, DepthTrack$_{miss}$, and VisEvent$_{miss}$ for comprehensive comparisons with state-of-the-art methods.
As reported in Tab.~\ref{table:rgbdt_comparison} and Tab.~\ref{tab-sota-miss}, AnyTrack achieves 68.5\% DP on RGBDT500$_{miss}$, 70.6\% PR on LasHeR$_{miss}$, 54.9\% F-score on DepthTrack$_{miss}$, and 54.0\% SR on VisEvent$_{miss}$, significantly outperforming other trackers.
These results validate that our method can handle different cases of missing modalities, enabling flexible modality adaptation.

\subsection{Ablation Study}

We validate the effect of key components and hyper-parameters through systematic ablations, with detailed analysis provided below.

\textbf{Effect of key modules.}
Tab.~\ref{tab:ablation} examines the effect of removing individual modules, reporting DP on RGBDT500, PR on LasHeR, PR on DepthTrack, and PR on VisEvent.
Removing MIM results in consistent performance degradation across all benchmarks, with drops of 2.3\% on RGBDT500 and 1.6\% on LasHeR.
This verifies that dynamic interaction and temporal aggregation are essential for fusing heterogeneous modalities.
Excluding CUM induces more severe declines, particularly a 3.4\% drop on VisEvent and 2.9\% on LasHeR.
This confirms the critical role of target-aware context modeling and position-adaptive guidance in maintaining discrimination.
The complementary nature of both modules is evident, as their integration yields the optimal performance across diverse tracking scenarios.

\textbf{Effect of interaction strategies.}
Tab.~\ref{tab:interaction} compares different interaction strategies on RGBDT500.
Graph Attention Networks (GAT) aggregate each modality via global average pooling and construct correlation graphs for feature interaction, incurring 101.6M parameters.
This mechanism loses spatial details for precise localization.
The direct concatenation along the token dimension followed by MLP fusion yields 75.3\% AUC, failing to model modality-specific patterns.
The bidirectional Mamba achieves 79.4\% DP, as sequential processing struggles with complex cross-modal interactions.
In contrast, our MIM achieves 79.2\% AUC with 90.6M parameters.
This validates that MIM effectively handles any modalities with high parameter efficiency and robust performance.

\begin{table}[t]
\centering
\caption{Ablation study of key modules.}
\label{tab:ablation}
\vspace{-3mm}
\renewcommand\arraystretch{0.85}{
\resizebox{\linewidth}{!}{
\begin{tabular}{cccccc}
\toprule
Method  & RGBDT500  & LasHeR & DepthTrack & VisEvent & $\Delta $ \\
\noalign{\hrule height 0.5pt}
\rowcolor[RGB]{254,236,221}
AnyTrack  & \textbf{86.1}  & \textbf{77.2} & \textbf{66.0} & \textbf{ 81.3} & $-$\\
w/o MIM  & 83.8  & 75.6 & 63.4  & 78.5 & -2.3\\
w/o CUM  &  82.2 & 74.3 &  62.9 & 77.9 & -3.3\\
\bottomrule
\end{tabular}
}
}
\vspace{-4mm}
\end{table}

\begin{table}[t]
\centering
    \caption{Comparison of interaction strategies.}
\label{tab:interaction}
\vspace{-3mm}
\renewcommand\arraystretch{0.85}{
\resizebox{\linewidth}{!}{
\begin{tabular}{cccccc}
\toprule
Module  & AUC $\uparrow$  & DP$\uparrow$  & Params $ \downarrow $ & FLOPs $ \downarrow $ & FPS $ \uparrow $\\
\midrule

GAT  & 76.0 & 81.5 & 101.6M  & 62.4G & 23 \\
MLP  &  75.3 & 80.5 & 90.8M  & 62.1G & 25 \\
Mamba  & 74.6 & 79.4 & 92.3M & 64.5G & 18\\
\rowcolor[RGB]{254,236,221}
MIM (Ours)  & \textbf{79.2}  & \textbf{86.1} & \textbf{90.6M} & \textbf{61.9G} & \textbf{26} \\
\noalign{\hrule height 0.8pt}
\end{tabular}
}
}
\vspace{-3mm}
\end{table}

\textbf{Effect of key components in MIM.}
As shown in the left part of Tab.~\ref{tab:variants}, experiments on RGBDT500 demonstrate that removing the noisy gating $\tilde{\mathbf{z}}_m$ degrades performance to 78.1\% AUC.
This verifies that injecting learnable noise into routing logits can encourage diverse expert activation during training.
Excluding the semantic expert $\mathcal{E}_s$ induces a more substantial decline to 84.2\% DP, confirming that capturing cross-modal shared semantics is critical for aligning heterogeneous features and maintaining robustness across diverse modalities.
Furthermore, without the balancing loss $\mathcal{L}_{\text{balance}}$ reduces accuracy to 78.5\% AUC and 85.4\% DP.
This indicates that regularizing expert routing to ensure a balanced utilization contributes to the training stability and prevents routing collapse.

\begin{table}[t]
\centering
\caption{Ablation study on MIM and CUM.}
\label{tab:variants}
\vspace{-3mm}
\newcolumntype{C}[1]{>{\centering\arraybackslash}m{#1}}
\renewcommand\arraystretch{0.85}{
\resizebox{\linewidth}{!}{
\begin{tabular}{C{1.8cm}C{0.9cm}C{0.9cm}|C{1.8cm}C{0.9cm}C{0.9cm}}
\toprule
\rule{0pt}{2.25ex}Module & \rule{0pt}{2.25ex}AUC$\uparrow$ & \rule{0pt}{2.25ex}DP$\uparrow$ & \rule{0pt}{2.25ex}Module & \rule{0pt}{2.25ex}AUC$\uparrow$ & \rule{0pt}{2.25ex}DP$\uparrow$ \\
\midrule
\cellcolor[RGB]{254,236,221}MIM & \cellcolor[RGB]{254,236,221}\textbf{79.2} & \cellcolor[RGB]{254,236,221}\textbf{86.1} & \cellcolor[RGB]{254,236,221}CUM & \cellcolor[RGB]{254,236,221}\textbf{79.2} & \cellcolor[RGB]{254,236,221}\textbf{86.1}  \\
w/o $\tilde{\mathbf{z}}_m$ & 78.1 & 85.0 &
w/o $\mathcal{Q}$ & 77.5 & 84.6 \\
w/o $\mathcal{E}_s$ & 77.4 & 84.2 & w/o $\mathbf{D}^t$  & 78.1 & 84.9 \\
w/o $\mathcal{L}_{\text{balance}}$ & 78.5 & 85.4 & w/o $\mathcal{V}$  & 78.2 & 85.3 \\
\bottomrule
\end{tabular}
}
}
\vspace{-3mm}
\end{table}

\textbf{Effect of key components in CUM.}
As shown in the right part of Tab.~\ref{tab:variants}, experiments on RGBDT500 demonstrate that eliminating the mask memory $\mathcal{Q}$ degrades performance by 1.7\% AUC.
This confirms that maintaining historical mask representations is essential for propagating spatial context through the asymmetric bidirectional attention.
Replacing the position-adaptive Gaussian pseudo-masks $\mathbf{D}^t$ with binary hard masks drops 1.2\% DP accuracy.
Furthermore, adopting static Gaussian variances instead of the dynamic variance estimation network $\mathcal{V}$ reduces performance by 1.0\% AUC.
This verifies that adapting spatial priors to target-specific scales is critical for maintaining precise localization.

\subsection{Visualization analysis}

\begin{figure}[t]
\centering
\includegraphics[width=\linewidth]{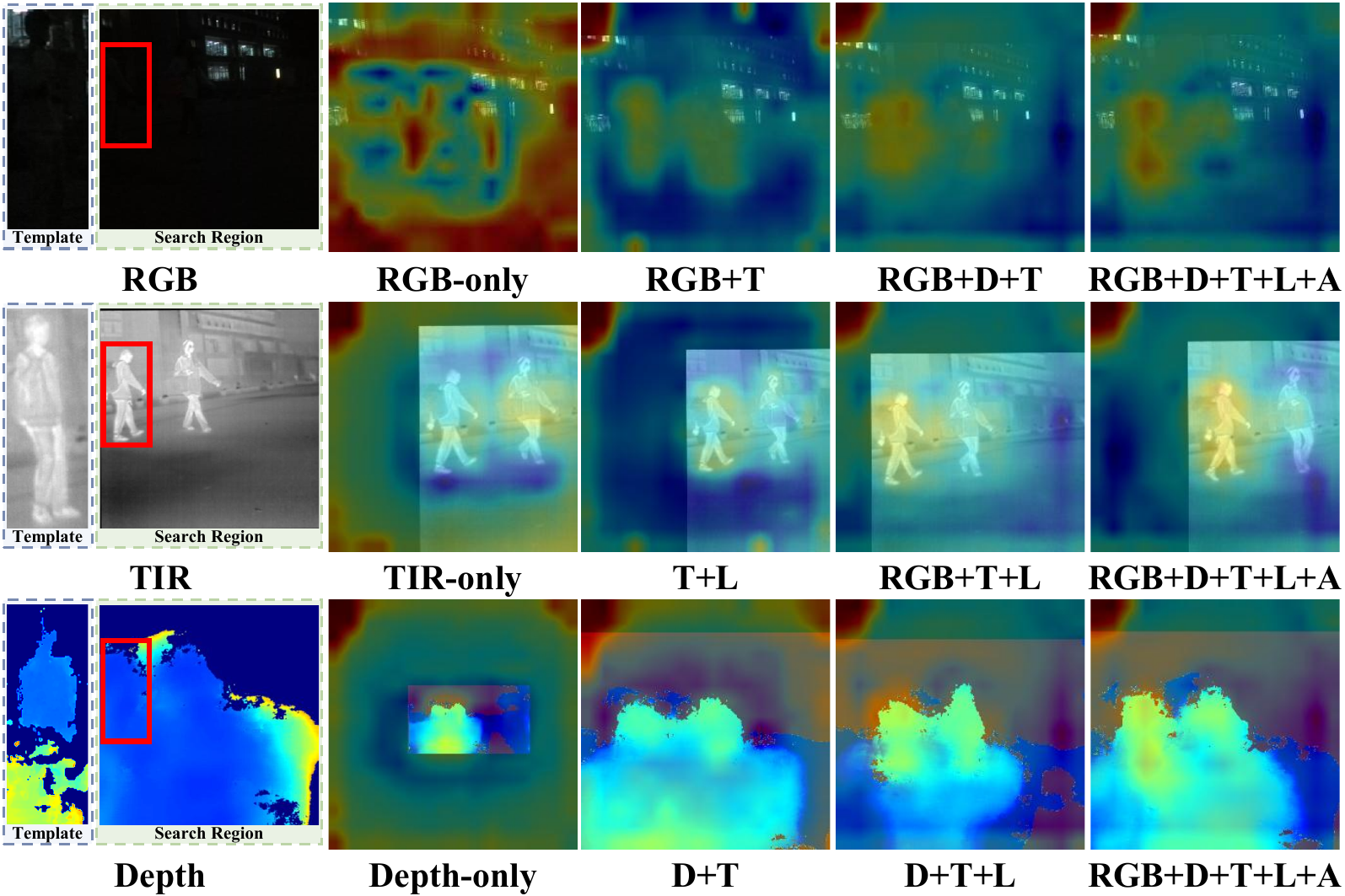}
\vspace{-5mm}
\caption{Visualization of attention maps.}
\label{fig:vis}
\vspace{-5mm}
\end{figure}

Fig.~\ref{fig:vis} presents attention map visualizations.
Attention maps reveal the progressive refinement from single-modal to multi-modal configurations.
With individual modalities, the model exhibits diluted responses due to limited feature discrimination.
Any multi-modal combinations progressively concentrate attention on target regions, achieving precise localization.
This validates that AnyTrack achieves robust tracking through dynamic interaction among any modalities, adaptively leveraging complementary cues.

\section{Conclusion}

In this work, we propose AnyTrack, a unified framework for visual object tracking with any modalities.
To handle any modality combinations within a single model, we design a Modality-aware Interaction Module (MIM) that dynamically bridges modality discrepancies and aggregates temporal cues to maintain spatio-temporal consistency.
Furthermore, we introduce a Context Understanding Module (CUM) that establishes spatial correspondence between visual features and target locations via global-local prompts, enabling precise target-aware context modeling.
To facilitate comprehensive training and evaluation, we extend existing multi-modal tracking benchmarks with additional modalities including grayscale images, language descriptions, and audio clips.
Extensive experiments on four tracking benchmarks validate the effectiveness and flexibility of our method in both complete and missing modality scenarios.


\bibliographystyle{ACM-Reference-Format}
\balance
\bibliography{sample-base}

\end{document}